\documentclass[acmtog]{acmart}

\usepackage{booktabs} 
\usepackage{amsmath,amssymb}

\setcitestyle{square}

\long\def\ignorethis#1{}

\usepackage{amsmath}
\usepackage{url}

\usepackage{xfrac}
\usepackage{newtxmath}
\usepackage{amsfonts,amsthm,amssymb,mathrsfs}

\usepackage{calrsfs}
\DeclareMathAlphabet{\pazocal}{OMS}{zplm}{m}{n}

\usepackage{booktabs}

\usepackage{soul}
\soulregister\ref{7}
\soulregister\cite{7}
\soulregister\refFig{7}

\usepackage{color}
\definecolor{gray}{rgb}{0.35,0.35,0.35}
\definecolor{blue}{rgb}{0,0,1}
\definecolor{white}{rgb}{1,1,1}

\usepackage{booktabs}

\usepackage{epsfig}
\usepackage{epstopdf}
\usepackage{subfig}
\usepackage{multirow}
\usepackage{array}

\usepackage{xcolor}

\newbox\jsavebox
\newcommand{\jsubfig}[2]{%
	\sbox\jsavebox{#1}%
	\parbox[t]{\wd\jsavebox}{\centering\usebox\jsavebox\\#2}%
	}

\newcommand{\Norm       } [1] {{\left\| #1 \right\|}}
\newcommand{\Pnorm      } [2] {\Norm{#1}_{#2}}

\setcopyright{none}

\hypersetup{draft}
\begin{document}
\title{Clustering-driven Deep Embedding with Pairwise Constraints} 
\author{Sharon Fogel}
\affiliation{
  \institution{Tel-Aviv University}
}

\author{Hadar Averbuch-Elor}
\affiliation{%
  \institution{Tel-Aviv University}
}

\author{Jacob Goldberger}
\affiliation{
  \institution{Bar-Ilan University}
}

\author{Daniel Cohen-Or}
\affiliation{
  \institution{Tel-Aviv University}
}

\renewcommand\shortauthors{Fogel et al.}

\begin{abstract}
Recently, there has been increasing interest to leverage the competence of neural networks to analyze data. In particular, new clustering methods that employ deep embeddings have been presented.
In this paper, we depart from centroid-based models and suggest a new framework, called Clustering-driven deep embedding with PAirwise Constraints (CPAC), for non-parametric clustering using a neural network. We present a clustering-driven embedding based on a Siamese network that encourages pairs of data points to output similar representations in the latent space. Our pair-based model 
allows augmenting the information with labeled pairs to constitute a semi-supervised framework. Our approach is based on analyzing the losses associated with each pair to refine the set of constraints. We show that clustering performance increases when using this scheme, even with a limited amount of user queries.
We demonstrate how our architecture is adapted for various types of data and present the first deep framework to cluster 3D shapes. 
\end{abstract}
\begin{teaserfigure}
\includegraphics[width=\textwidth]{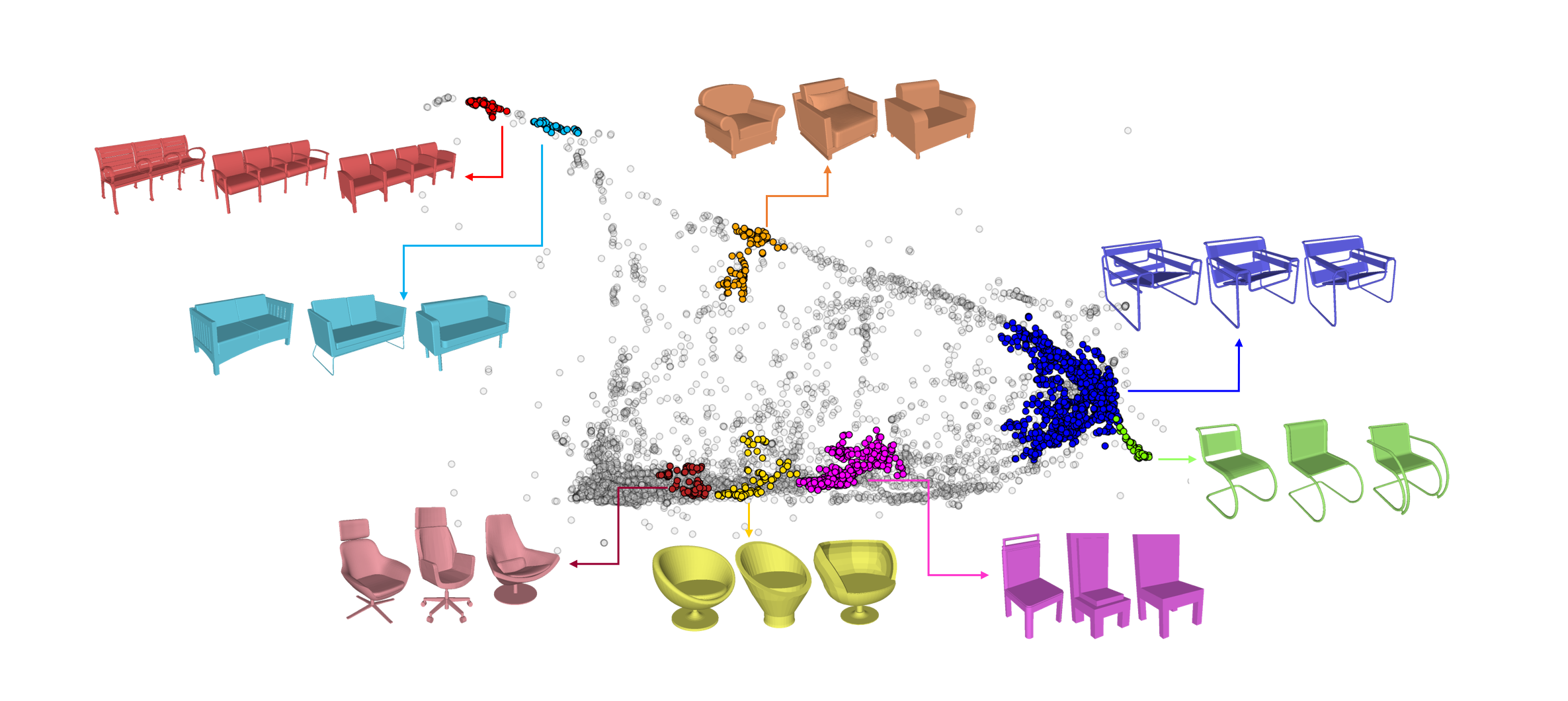}
\caption{Employing deep embeddings for clustering 3D shapes. Above, we use PCA to visualize the output embedding of point clouds of chairs. We also highlight (in unique colors) a few random clusters and display a few representative chairs from these clusters.
}
\label{fig:teaser}
\end{teaserfigure}

\maketitle

\section{Introduction}

Autoencoders provide means to analyze data without supervision. Autoencoders based on deep neural networks include non-linear neurons which significantly strengthen the power of the analysis. The key idea is that the encoders project the data into an embedding latent space, where the $L_2$ proximity among the projected elements better expresses their similarity. To further enhance the data proximity in the embedding space, the encoder can be encouraged to form tight clusters in the embedding space. Xie et al. \shortcite{xie2016unsupervised} have presented an unsupervised embedding driven by a centroid-based clustering. They have shown that their deep embedding leads to better clustering of the data. 
More advanced clustering-driven embedding techniques have been recently presented \cite{yang2016joint,dizaji2017deep}. These techniques are all centroid-based and parametric, in the sense that the number of clusters is known a-priori.

In this paper, we present a clustering-driven embedding technique that allows semi-supervision. The idea is to depart from centroid-based methods and use pairwise constraints to drive the clustering. Most, or all the constraints, can be learned with no supervision, while possibly a small portion of the data is supervised. More specifically, we adopt robust continuous clustering (RCC) \cite{shah2017robust} as a driving mechanism to encourage a tight clustering of the embedded data.


\begin{figure}
\includegraphics[width=\columnwidth]{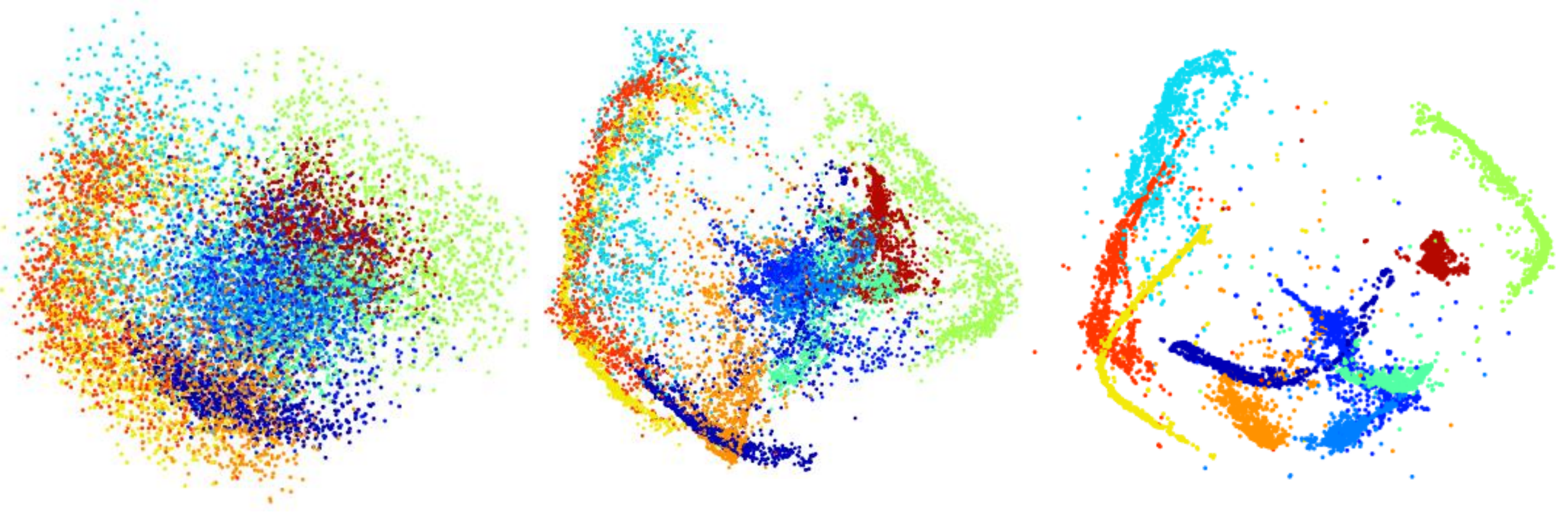}
(a) \hspace{70pt} (b)\hspace{70pt} (c) 
\caption{Clustering the USPS dataset with pairwise constraints. Above, we visualize the encoded data (a) after the initial training of the autoencoder, (b) after training with the additional clustering loss, and finally, (c) after training with 1k constraints labeled by the user. The visualizations were obtained by running PCA on the encoded data.   }
\label{fig:pca_usps}
\end{figure}

The idea is to extract pairwise constraints using a mutual $k$-nearest neighbors analysis, and use these pairs as must-link constraints. With no supervision, the set of constraints is imperfect and contains false positive pairs on one hand. Our technique allows removing false positive pairs and strengthening true positive pairs actively by a user. 
We present an approach that analyzes the losses associated with the pairs to form a set of false positive candidates. See Figure \ref{fig:pca_usps}(b)-(c) for a visualization of the distribution of the data with and without the active removal of a small portion of false positive constraints (less than $0.01\%$ out of all possible pairs) .

\begin{figure*}
\includegraphics[width=\textwidth]{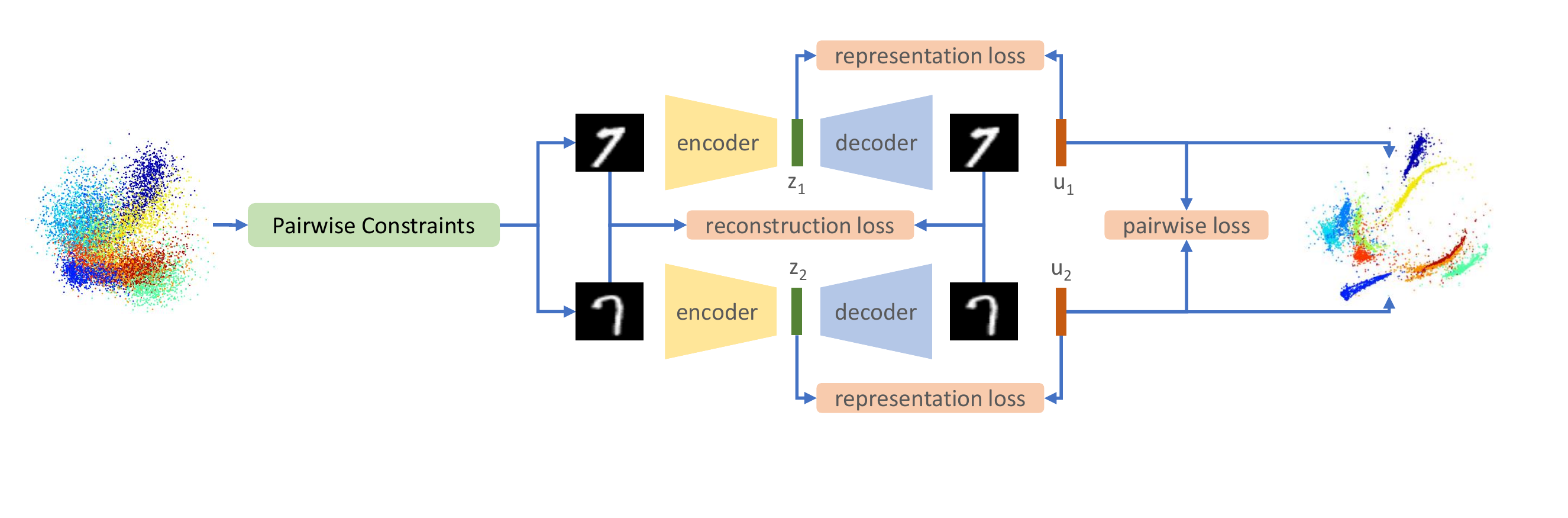}
\caption{Overview of our method. Constrained pairs are fed into our network which jointly optimizes their unary loss, constructed from a representation loss and a reconstruction loss, as well as their pairwise loss. These pairs are computed automatically from the data and are possibly augmented with pairs of labeled data. }
\label{fig:overview}
\end{figure*}

Given the pairwise constraints which are computed automatically from the data and are possibly augmented with pairs of labeled data, we train a Siamese network to output a similar representation among the pairs (using a pairwise loss) while remaining loyal to the input data (using a reconstruction loss). 
To increase the flexibility of our system, we reformulate the loss function as a constrained optimization problem and train our network in an alternating fashion using an ADMM technique \cite{boyd2011distributed}. Our constrained formation allows to disentangle between the clustering and the representation loss, while remaining faithful to the original problem formulation.
Our network and the different losses are illustrated in Figure \ref{fig:overview}. 

We apply our semi-supervised clustering technique to solve the challenging problem of sub-categorizing a class of shapes \cite{kleiman2015shed,fish2016structure}. 
There are many repositories of three-dimensional shapes, which are categorized into classes, however, sub-categorization is not provided. For example, Shapenet
\cite{chang2015shapenet} 
contains about 51,300 3D shapes represented by point clouds, classified into 55 classes. In Figure \ref{fig:teaser}, we show the results of our network-based clustering on the class of chairs. The class of chairs is quite large and diverse, and contains a variety of chairs types. As can be observed, the network embeds the chairs in a meaningful way, and clusters the sub-categories close together. It should be noted that the number of clusters is unknown a-priory. In Section \ref{sec:exp}, we provide more results.

We evaluate the performance of our method on a variety of datasets. First, we show that our technique leads to state-of-the-art deep clustering results and outperforms RCC and other non-parametric clustering methods (where the number of clusters is unknown). Furthermore, our results are comparable to deep embedding techniques that are centroid-based. 
Finally, we evaluate our semi-supervised setting, and show that the deep clustering performance is enhanced with the addition of pairwise constraints.

\section{Related Works}

Clustering is a fundamental data analysis tool, with abundant applications
in different fields of science. Thus, it has been an active topic of research for
the past few decades \cite{berkhin2006survey,xu2005survey}.
Clustering methods can be broadly categorized into parametric and non-parametric techniques, according to whether the parameter representing the number of clusters is provided as input or not.

Density based methods, such as DBSCAN \cite{ester1996density} and Mean-Shift \cite{cheng1995mean}, group together points that are packed together closely, without knowing the number of clustering a-priori. However, these methods tend to be sensitive to their input parameters, as shown, for example, in \cite{karypis1999chameleon}. More recent extensions attempt to overcome this issue and demonstrate a more robust solution \cite{comaniciu2001variable,campello2013density}. However, these techniques are still dependent on the input data, which may vary across datasets with varying distributions.

To cope with a wider set of scenarios, feature learning techniques have been used alongside the clustering methods, originally in a sequential manner. The input data is first transformed to a representation which is more cluster-friendly, and clustering is then applied on the obtained feature space \cite{ding2004k,tian2014learning}.

The RCC algorithm \cite{shah2017robust} suggested a new way to create an embedding of the data which is more suitable for clustering. The algorithm brings mutual KNN points in the original embedding closer together in the clustering embedding to create tighter clusters. 

The DEC \cite{xie2016unsupervised} algorithm coupled the learning and the clustering stages into one unified deep optimization scheme. Using a stacked autoencoder, a cluster-friendly embedding of the data is learnt simultaneously to a clustering of the data. Many extensions followed, boosting performance and adapting the network architecture. The JULE \cite{yang2016joint} method proposes to represent the data using a convolutional network and optimizes the embedded space using a recurrent framework. The method uses a loss function derived from the merging process of agglomerative clustering. The method demonstrates improved performance over various benchmarks, however, to use this technique, a large set of parameters must be tuned for each dataset. The DEPICT method \cite{dizaji2017deep} uses a convolutional autoencoder stacked with a soft-max layer. The soft-max layer is used as a discriminative clustering model which is trained using relative entropy minimization, as well as a regularization term to balance the frequency of cluster assignment. Thus, this technique is susceptible to datasets that contain clusters of different sizes. 
Similarly to these methods, we also jointly optimize the learning and clustering stages using a deep platform. However, \emph{all} previous deep techniques \emph{explicitly} assume the number of clusters is given, while we present a non-parametric approach inspired from the RCC algorithm \cite{shah2017robust}, avoiding such assumptions. 

\paragraph{Semi-supervised clustering} In semi-supervised clustering, some supervision is given on a (typically small) subset
of the data in the form of labeled data, or as pairs of
constraints. 
Several works have used the constraints to define a new distance matrix that warps the standard Euclidean metric into a Mahalanobis one \cite{xing2003distance,bar2003learning,davis2007information,hoi2010semi}. Others have used the pairwise constraints to change the embedding of the data \cite{wang2012active,asafi2013constraints}. In all of these approaches, the constraints are taken into account in the preprocessing of the data before the clustering algorithm and not as part of the clustering.
There have been numerous papers combining pairwise constraints into spectral clustering methods by making local changes to the affinity matrix according to the constraints \cite{kamvar2003spectral}. Some attempts have been made at improving this method by propagating the constraints themselves to nearby pairs \cite{sharma2010learning,he2012constrained,lu2010constrained}.
In \cite{hsu2015neural}, pairwise constraints are used for clustering in a siamese architecture. However, since the unlabeled data is not taken into account, a large amount of constraints is needed for clustering. 

Recent works have used partially labeled data for clustering using a deep architecture which is trained both on labeled and unlabeled data \cite{kingma2014semi,rasmus2015semi,dizaji2017deep,zhao1506stacked}. Contrarily to our method, these techniques use labeled points as opposed to labeled connections and also the number of classes in the dataset is known a priori.
Our method enables an effective semi-supervision scheme, where performance can be improved with limited user intervention. When such supervision is not available, our approach reduces to an unsupervised deep clustering technique. 
\section{Clustering with Pairwise Constraints}
In this work, we consider the problem of clustering a set of $n$ data points into an unknown number of clusters. Denote the data points by $X=[x_1,x_2,...,x_n]$. Rather than clustering the data points in their original space, we use a deep autoencoder to create a representation of the data in a latent space which is better suited for clustering.

Similarly to \cite{xie2016unsupervised}, our method operates in two stages. 
In the first step, we train a multilayer autoencoder and initialize the latent space using its learned representation layer, $Z=\textsc{Enc}(X)$. In the second step, we add a clustering loss on top of the reconstruction loss, and continue training the same autoencoder in an alternating fashion using an ADMM techinque \cite{boyd2011distributed}, which helps in decoupling the reconstruction and the clustering loss.

\subsection{Latent Space Initialization}
Recent clustering methods that utilize autoencoders, such as \cite{yang2016joint} and \cite{dizaji2017deep}, use convolutional autoencoders, as they have been proven beneficial, especially for visual data. However, to obtain a clustering technique which is applicable for any type of data, we use a linear autoencoder, similar to the one used in \cite{xie2016unsupervised}, with ReLU activations and dropout layers. 
Nonetheless, in Section \ref{sec:exp}, we demonstrate that our general architecture can easily be suited for specific types of datasets in two ways: (i) by complementing our architecture with a well-known pretrained net, such as VGG \cite{simonyan2014very} for images, or (ii) by using a different architecture for the autoencoder as we demonstrate for point cloud clustering.


\subsection{Optimization of the Clustering Embedding}
After training the $\textsc{Enc}-\textsc{Dec}$ network (the entire autoencoder net) on a reconstruction loss we have an initial non-linear mapping of the original data points $X$ to the latent representation $Z$. The latent representations is optimized such that pairs of data points which have a high probability of belonging to the same cluster will be drawn closer together, thus creating a representation which is more meaningful for clustering.

Our approach for optimizing the clustering embedding is inspired by the recent work of Shah et al. \cite{shah2017robust}. We adapt their driving mechanism to a deep learning framework. While many details, including heuristics for initializing the various components, remain similar, we leverage a deep neural network which enables a more complex representation of the data.

Following \cite{shah2017robust}, our loss function combines a clustering loss that encourages a formation of clusters in the data together with a reconstruction term that prevents the data from collapsing into one single cluster and keeps the latent representation meaningful.

More formally, let $\theta_e$ be the weights of the encoder net  with output $Z = \textsc{Enc}(X)$ and $\theta_d$ be the weights of the decoder net with output $X' = \textsc{Dec}(Z)$. We define $L_{cluster}(\theta_e)$ to be the clustering loss and $L_{rec}(\theta_e, \theta_d)$ to be the reconstruction loss.The objective is to optimize the sum of the two losses:
\begin{equation} 
\label{eq:loss_function}
\operatorname*{min}_{\theta_e, \theta_d}\dfrac{1}{\dim{(X)}} \sum_{i=1}^{N}{ \Pnorm{x'_i - x_i}{2}^2} + \dfrac{\lambda}{\dim{(Z)}} \sum_{(p,q)\in \varepsilon}{ w_{p,q}\rho_2 (\Pnorm{z_p - z_q}{2}^2)}.\\ 
\end{equation}
The first term is a reconstruction loss of the autoencoder. 
The second term is a distance measurement between the clustering representation of pairs of data points. Pairs are assembled from a connectivity graph $\varepsilon$, which we construct using a mutual KNN (MKNN) connectivity between data points in the initial space $X$. The usage of MKNN rather than a KNN connectivity creates a more conservative graph, where only data points that are mutually close are connected.

The weights $w_{p,q}$ balance the contribution of each data point to the loss. Each weight is determined according to:
\begin{equation}
w_(p,q) = \dfrac{\dfrac{1}{N}\sum_i{n_i}}{\sqrt[2]{n_{p}n_{q}}}.
\end{equation}

We use the Geman-McClure norm as the penalty function $\rho(y)$: \begin{equation}
\label{eq:Geman-McClure}
\rho_i(y)=\dfrac{\mu_i y^2}{\mu_i + y^2} \hspace{1cm} i=1,2,
\end{equation}
where the parameters $\mu_1$ and $\mu_2$ used in $\rho_1$ and $\rho_2$, respectively, are initialized to be larger than the distances between data points in order to avoid non-convexity, and are gradually minimized every few epochs to encourage a smaller influence of the outlier connections on the loss function.

The normalizing parameter $\lambda$ is computed according to:
\begin{equation}
\label{eq:lamda_calc}
\lambda = \dfrac{\Pnorm{Z}{2}}{\Pnorm{D-R}{2}},
\end{equation}
where $\Pnorm{\cdot}{2}$ denotes the spectral norm, $Z$ is the original data representation in the latent space, $R$ is a sparse matrix of $N\times N$ with the values $w_{i,j}$ at places $(i,j)$ and $D$ is a diagonal matrix with the values $\sum_{j=1}^{N}{w_{i,j}}$ along the diagonal. The parameter $\lambda$ is calculated once at the beginning of the run.

We normalize the reconstruction loss by $\dim{(X)}$, the dimension of the input data $X$ and the clustering loss by $\dim{(Z)}$, the dimensionality of the embedding space.

To add flexibility to our optimization scheme, we can rewrite the loss function described in Eq. \ref{eq:loss_function} as a constrained optimization problem:
\begin{align}
\label{eq:ADMM loss}
\begin{split} 
&\operatorname*{min}_{\theta_e, \theta_d, U}\dfrac{1}{\dim{(X)}} \sum_{i=1}^{N}{ \Pnorm{x'_i - x_i}{2}^2} + \dfrac{\lambda}{\dim{(Z)}} \sum_{(p,q)\in \varepsilon}{ w_{p,q}\rho_2 (\Pnorm{u_p - u_q}{2}^2)}. \hspace{0.5cm} \\
& s.t \hspace{0.5cm} u_i=z_i \hspace{0.1cm} \forall i\\ 
\end{split}
\end{align}
The new formulation contains an additional representation $U$. It is equivalent to the original loss function, since the constraint forces the two representations to be identical. However, it allows for a decoupling of the larger complicated loss function introduced in Eq. \ref{eq:loss_function} into separate more manageable optimization problems on the clustering and reconstruction loss. The constrained problem can be solved using an ADMM technique \cite{boyd2011distributed} which alternates between minimizing the loss function with respect to $U$ and the autoencoder weights $\theta_e,\theta_d$ and updating the Lagrange multiplier. The new optimization method can be described by the following steps:
\begin{align}
\label{eq:ADMM method}
\begin{split}
1.\hspace{0.5cm} &U = \operatorname*{argmin}_{U} \dfrac{\lambda}{\dim{(Z)}} \sum_{(p,q)\in \varepsilon}{ w_{p,q}\rho_2 (\Pnorm{u_p - u_q}{2}^2)} +\\
&\dfrac{1}{\dim{(Z)}} \sum_{i=1}^{N}{ \rho_1 ( \Pnorm{z_i - u_i}{2}^2 )} + \vartheta(Z-U) \\
2.\hspace{0.5cm}  &\theta_e, \theta_d = \operatorname*{argmin}_{\theta_e, \theta_d} \dfrac{1}{\dim{(X)}} \sum_{i=1}^{N}{ \Pnorm{x'_i - x_i}{2}^2} +\\
&\dfrac{1}{\dim{(Z)}} \sum_{i=1}^{N}{ \rho_1 ( \Pnorm{z_i - u_i}{2}^2 )} + \vartheta(Z-U) \\
3.\hspace{0.5cm}  &\vartheta = \vartheta + a(Z-U)
\end{split}
\end{align}
where $\vartheta$ is the Lagrange multiplier.

Our loss function in the first stage of Eq. \ref{eq:ADMM method} includes both a binary term and a unary term. To avoid optimizing each one of them separately, we train our network on the entire loss function using pairs of points as input to the network. The unary term is computed on both points and additional weights are added to better balance between data points with a varying connectivity. 
The weight of the unary loss term is $w_i=1 / N_i$, where $N_i$ is the total number of connections a point $x_i$ has. Since each clustering representation $u_i$ is trained only in a subset of the batches, we assign a higher learning rate to $U$.

\subsubsection{Final Clustering}
The final clustering is computed according to the connectivity graph $\varepsilon$, which is created at the initialization of the clustering algorithm. Pairs which are within a distance smaller than the defined threshold remain connected, and pairs with larger distances are disconnected, thus creating a new connectivity graph $\psi$. The output clusters are the connected components of $\psi$.

\section{Semi-Supervised Clustering}
\label{sec:semi_method}

Our method can be further extended to a semi-supervised clustering technique by simply strengthening or removing manually labeled pairs to the pairwise constraints. 
Rather than selecting the labeled pairs randomly, we suggest a mechanism to perform a \emph{smart} pair selection, using either must-link or cannot-link constraints. To incorporate the new constraints, the loss term defined in Eq. \ref{eq:loss_function} is adjusted accordingly. In other words, edges on the connectivity graph $\epsilon$ are either strengthened or removed, according to the manually labeled pairs.    

\begin{figure*}[t]
\includegraphics[width=\textwidth]{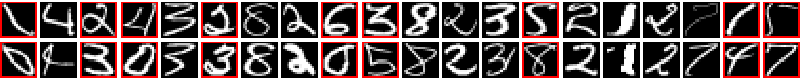}
\includegraphics[width=\textwidth]{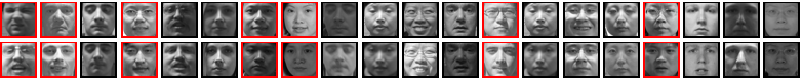}
\caption{Cannot-link constraints in the USPS and FRGC datasets. Pairs exhibiting the highest clustering loss value at initialization are depicted from left to right. The cannot-link constraints (visualized with a red frame) are selected according to the ground truth labels.}
\label{fig:high_loss_pairs}
\end{figure*}

To strengthen or remove existing constraints from the connectivity graph $\epsilon$, we examine ``hard'' pairs, i.e., pairs that did not perform well by the network. Intuitively, these pairs have a higher chance of being false-negative connections, which we would like to eliminate from the connectivity graph $\epsilon$. We sort the pairwise connections according to the clustering loss defined in the second term of Eq. \ref{eq:loss_function} after the initial autoencoder training. Given pairs with a high clustering loss, the user selects which pairs should not be incorporated into the formulation (as they contain points that belong to two different clusters). These labeled pairs constitute our cannot-link constraints (visualized with a red frame in Figure \ref{fig:high_loss_pairs}). Pairs that should belong to the same cluster according to the user constitute our must-link constraints, and are set with a weight equal to the maximum connection in the connectivity graph $\epsilon$. See Figure \ref{fig:high_loss_pairs} for a visualization of the annotated pairs on two different datasets.

\section{Results and Evaluation}
\label{sec:exp}

\subsection{Implementation Details}
\label{sec:imp}
The implementation of our method is available at: \url{https://github.com/sharonFogel/CPAC}. Throughout the evaluation of our method in the experiments described below, the method was run in 
the following way: we used autoencoder dimensions of $\dim{(X)}-500-500-2000-10$. 50 epochs of layerwise training were used for pre-training of the autoencoder, followed by 50 additional epochs for fine-tuning the entire autoencoder.
We used an Adam optimizer for training the network, with a learning rate of $0.0001$ and $\beta=(0.9,0.999)$. During the layerwise training, a dropout of 20\% was used after each layer, except for the last layer of the decoder.
The parameters $\delta_1$, $\delta_2$ are computed prior to training the network on the clustering loss. The parameter $\delta_1$ is set to be twice the average distance of the encoded data from the mean of the encoded data. The parameter $\delta_2$ is set to be the average distance between the nearest 1\% of the MKNN connections (with a cut-off of 250 nearest neighbors). We initialize $\mu_1$ and $\mu_2$ to be $8\cdot\delta_1$ and $3\cdot \max{\Pnorm{u_p - u_q}{2}^2}$. During the optimization of the clustering embedding, the parameters $\mu_1$ and $\mu_2$ are divided by two every constant number of epochs until reaching the lower bound $\delta_1$ and $\delta_2$. The number of epochs between each update is set according to the percentage of MKNN connections in the data (out of all the possible $N\times N$ connections). It is set to 60 for datasets with less than $0.2\%$ connections and 10 for datasets with a higher percent of MKNN connections. Each epoch the optimization is alternated between $U$ and $f_\theta$, and after one optimization cycle the Lagrange multiplier is updated. An RMS optimizer is used for training. We set a different learning rate for the autoencoder weights and for the clustering representation $U$. The autoencoder learning rate is set to 0.0001 and the learning rate of $U$ is set to 0.04 in all of the experiments.
The threshold used in the final clustering is set to be the average distance between the nearest 1\% out of the MKNN connections.
\subsubsection{Using Pretrained VGG Features} 
To obtain a clustering technique which is applicable to any type of data, we purposely avoid convolutional layers. However, we demonstrate that our general architecture can easily be complemented 
with a pretrained VGG net to boost our scores on image data.
We use the pretrained VGG19 model on Pytorch. Before passing the images through the network, we resize them to $224\times224$. We feed the output of the convolutional layers of the VGG to our autoencoder. The learning rate for $U$ in the VGG experiments is $0.1$. The number of epochs between the $\mu_1$ and $\mu_2$ update is set according to the percentage of MKNN connections in the data. It is set to 60 for datasets with less than $0.2\%$ connections, and to 40 for datasets with a higher percentage of MKNN connections.

\subsection{Unsupervised Clustering Evaluation}
\subsubsection{Alternative Algorithms}
We are not aware of any other deep clustering method which is non-parametric.
Therefore, we compare our clustering technique both to non-parametric well-known clustering algorithms and to state-of-the-art parametric clustering algorithms using deep learning. The non-parametric methods used are Affinity Propagation (AP) \cite{frey2007clustering}, Hierarchical-DBSCAN (HDB) \cite{campello2013density}, RCC and RCC-DR \cite{shah2017robust}. The parametric methods we compare to are DEC \cite{xie2016unsupervised}, JULE \cite{yang2016joint} and DEPICT \cite{dizaji2017deep}. 

\subsubsection{Datasets}

\begin{table}
  \begin{center}
  \setlength{\tabcolsep}{0.5em} 
    \begin{tabular}{ lccccc }
      \toprule
       	Dataset		& USPS			& FRGC			& YTF			& CMU-PIE			& REUTERS \\
        \midrule
       	\#samples	& 11,000		& 2462			& 10,000		& 2856				& 10,000 \\
       	\#classes	& 10			& 20			& 41			& 68				& 4 \\
        size  & 16$\times$16	& 32$\times$32	& 55$\times$55	& 32$\times$32		& 2000 \\
      \bottomrule
    \end{tabular}
  \end{center}
  \caption{Parameters of the datasets used for evaluation.}
  \label{tbl:datasets}
\end{table}

To evaluate our method, we use datasets of various categories. We use the USPS dataset that contains handwritten digits, three datasets of faces - FRGC, YouTube Faces (YTF) \cite{wolf2011face}, and CMU-PIE \cite{sim2002cmu}, and a subset of the REUTERS dataset which contains 10000 samples of English news stories labeled into 4 categories \cite{lewis2004rcv1}. The dataset parameters are described in Table \ref{tbl:datasets}. Following \cite{yang2016joint}, as pre-processing for the YTF and FRGC datasets, we crop and resize the faces to a constant size. In FRGC, we use the same 20 randomly chosen subjects as in \cite{yang2016joint}. In YTF, we use the first 41 subjects, sorted according to their names in alphabetical order.

\subsubsection{Quantitative Evaluation}
\begin{table*}[]
\centering
\setlength{\tabcolsep}{0.5em}
\begin{tabular}{l|cc|cc|cc|cc|cc}
           & \multicolumn{2}{c|}{USPS}                                                     & \multicolumn{2}{c|}{FRGC}                                                    & \multicolumn{2}{c|}{YTF}                                                      & \multicolumn{2}{c|}{CMU-PIE}                                                  & \multicolumn{2}{c}{REUTERS}                                                       \\ \cline{2-11} 
           & NMI                                   & ACC                                   & NMI                                   & ACC                                  & NMI                                   & ACC                                   & NMI                                   & ACC                                   & NMI                                   & ACC                                   \\ \hline
DEC        & 0.59                                 & 0.62                                 & 0.50                                 & 0.38                                & 0.45                                 & 0.37                                 & 0.92                                 & 0.80                                 & \textbf{0.72}  & {\textbf{0.57}}                                \\
JULE-SF    & 0.86                                 & 0.92                                 & 0.57                                 & 0.46                                & \textbf{0.85}                                 & \textbf{0.68}                                 & 0.98                                 & 0.98                                  & -                                     & -                                     \\
JULE-RF    & 0.91                                 & 0.95                                  & 0.57                                 & 0.46                                & \textbf{0.85}                                 & \textbf{0.68}                                 & {\textbf{1.00}}     & {\textbf{1.00}}     & -                                     & -                                     \\
DEPICT     & {\textbf{0.93}} & {\textbf{0.96}} & \textbf{0.61}                                  & \textbf{0.47}                                 & 0.80                                 & 0.62                                 & 0.97                                 & 0.88                                 & -                                     & -                                     \\
\hline
AP         & 0.54                                 & 0.04                                 & 0.58                                 & 0.20                                & 0.78                                 & 0.25                                 & 0.67                                 & 0.28                                 & 0.37                                 & 0.01                                 \\
HDB        & 0.37                                 & 0.25                                  & 0.55                                 & 0.23                                & 0.84                                 & {\textbf{0.60}} & 0.70                                 & 0.24                                 & 0.30                                   & {\textbf{0.22}}                                 \\
RCC        & 0.78                                 & 0.59                                 & 0.68                                 & 0.49                                & 0.84                                 & 0.46                                 & 0.88                                 & 0.62                                 & 0.36                                 & 0.02                                  \\
RCC-DR     & 0.76                                 & 0.52                                  & 0.60                                 & 0.43                                 &{\textbf{0.87}} & 0.44                                 & 0.51                                 & 0.23                                 & 0.37                                 & 0.02                                  \\
CPAC     & 0.77                                 & 0.58                                 & 0.74                                 & 0.48                                 & 0.86                                  & 0.51                                 & 0.87                                 & 0.63                                 & {\textbf{0.40}}                                 & 0.18                                  \\
CPAC-VGG   & {\textbf{0.92}}                                  & {\textbf{0.87}}                                  & {\textbf{0.80}}                                  & {\textbf{0.54}}                                  & 0.86                                  & 0.52                                 & {\textbf{0.90}}                                 & {\textbf{0.77}}                                  & -                                     & -                                    
\end{tabular}
\caption{Clustering evaluation on parametric (top rows) and non-parametric (bottom row) techniques. For each category, we emphasize the best results in bold. As the table illustrates, our non-parametric technique (bottom two rows) yields scores which are comparable to the parametric ones.}
\label{tbl:unsupervised_results_admm}
\end{table*}
To perform a quantitative evaluation of our algorithm, we use two well-known clustering measurements - normalized mutual information (NMI) and accuracy (ACC). 
NMI is defined according to:
\begin{equation}
\label{eq:NMI}
NMI(c,c') = \dfrac{I(c;c')}{max(H(c),H(c'))}
\end{equation}
where $c$ is the ground truth classification and $c'$ the predicted cluster. $I$ denotes the mutual information between $c$ and $c'$ and $H$ denotes their entropy. NMI has a tendency to favor fine-grained partitions. For this reason, we also report the ACC scores, which are defined according to:
\begin{equation}
\label{eq:ACC}
ACC(c,c') = \max_m{\dfrac{\sum_{i=1}^{N}{\textbf{1}\{c_i=m(c'_i)\}}}{n}}
\end{equation}
where $m$ ranges over all possible mappings between clusters and labels.

Table \ref{tbl:unsupervised_results_admm} contains the full quantitative analysis. 
For the alternative techniques, we report the scores which were reported in the original papers, and when these are not available, we run their publicly available code on our examined datasets. Methods that use convolutional networks cannot cluster non-visual data such as the REUTERS dataset, and therefore we use a dashed line (-) when clustering is not possible.

We report two variants of our algorithm: CPAC and CPAC-VGG. CPAC is our general framework, and CPAC-VGG is our framework complemented with a pretrained VGG (the full implementation details are provided in Section \ref{sec:imp} above).
As Table \ref{tbl:unsupervised_results_admm} illustrates, our method obtains state-of-the-art results and is comparable even to state-of-the-art parametric techniques, where the number of clusters in the data is provided. 

\begin{table*}[]
\centering
\resizebox{\textwidth}{!}{%
\begin{tabular}{c|l|cc|cc|cc|cc|cc}
\multicolumn{1}{l|}{}                                                                               & \multirow{2}{*}{Clustering Method} & \multicolumn{2}{c|}{USPS} & \multicolumn{2}{c|}{FRGC} & \multicolumn{2}{c|}{YTF} & \multicolumn{2}{c|}{CMU-PIE} & \multicolumn{2}{c}{REUTERS} \\ \cline{3-12} 
\multicolumn{1}{l|}{}                                                                               &                                    & NMI        & ACC          & NMI         & ACC         & NMI        & ACC         & NMI           & ACC          & NMI          & ACC           \\ \hline
\multirow{4}{*}{\begin{tabular}[c]{@{}c@{}}Clustering on\\  Initial Deep \\ Embedding\end{tabular}} & K-means                            & 0.61       & 0.57         & 0.44        & 0.35        & 0.77       & 0.58        & 0.62          & 0.31         & 0.72         & 0.52          \\
                                                                                                    & agglomerative                      & 0.6        & 0.55         & 0.5         & 0.37        & 0.8        & 0.62        & 0.64          & 0.33         & 0.43         & 0.64          \\
                                                                                                     & AP                                 & 0.57       & 0.06         & 0.6         & 0.26        & 0.81       & 0.32        & 0.71          & 0.31         & 0.4          & 3e-3         \\
                                                                                                    & HDBSCAN                            & 0.39       & 0.31         & 0.47        & 0.32        & 0.81       & 0.61        & 0.6           & 0.25         & 0.3          & 0.23          \\ \hline
\multirow{4}{*}{\begin{tabular}[c]{@{}c@{}}Clustering on \\ Final Deep \\ Embedding\end{tabular}}   & K-means                            & 0.69       & 0.60          & 0.59        & 0.45         & 0.83       & 0.64        & 0.87          & 0.65         & 0.48         & 0.72          \\
                                                                                                    & agglomerative                      & 0.69       & 0.59         & 0.58        & 0.44        & 0.83       & 0.65        & 0.85          & 0.65         & 0.44         & 0.75          \\
                                                                                                    & AP                                 & 0.79       & 0.58         & 0.73         & 0.50        & 0.85       & 0.53        & 0.87          & 0.65         & 0.39         & 0.18          \\
                                                                                                    & HDBSCAN                            & 0.79       & 0.58         & 0.72        & 0.51        & 0.87       & 0.53        & 0.87          & 0.66         & 0.38         & 0.18         
\end{tabular}%
}
\caption{Results of different clustering methods on the initial deep embedding and on the final deep embedding after training with the clustering loss. There is a clear improvement in the results of the different methods proving that the method creates a better embedding for clustering.}
\label{tbl:clust_embedded_space}
\end{table*}
We evaluate the deep embedding learned by our algorithm by showing the results of other clustering algorithms on the initial and final deep embedding in \ref{tbl:clust_embedded_space}. There is a significant improvement in the clustering results across the different methods and different datasets, especially across the non-parametric clustering methods. 


\subsection{Ablation Experiment}
\begin{table*}[]
\centering
\begin{tabular}{l|cc|cc|cc|cc|cc}
\multirow{2}{*}{\begin{tabular}[c]{@{}l@{}}Loss and Optimization\\  Method\end{tabular}} & \multicolumn{2}{c|}{USPS} & \multicolumn{2}{c|}{FRGC} & \multicolumn{2}{c|}{YTF} & \multicolumn{2}{c|}{CMU-PIE} & \multicolumn{2}{c}{REUTERS} \\ \cline{2-11} 
                                                                                         & NMI         & ACC        & NMI         & ACC         & NMI         & ACC        & NMI           & ACC          & NMI          & ACC           \\ \hline 
(i) $L_{rec}+L_{clust}$ (best results*)            & 0.50         & 0.04       & 0.66        & 0.38        & 0.84        & 0.58       & 0.55          & 0.25         & 0.10          & 0.37          \\
(ii) $L_{clust}$ (best results*)                    & 0.50         & 0.04       & 0.66        & 0.38        & 0.84        & 0.58       & 0.54          & 0.23         & 0.37         & 3e-3        \\
(iii) $L_{rec}+L_{rep}+L_{clust}$              & 0.77            & 0.58           & 0.74            & 0.48            & 0.86            & 0.51           & 0.87              & 0.63             & 0.40             & 0.18             
\end{tabular}
\caption{Clustering results with different optimization methods. For (i) and (ii), we try various parameter settings (using a grid of possible parameter configurations) and report 
the best results. }
\label{tbl:optimization_methods}
\end{table*}
To better understand the effect of each of the loss on the optimization process we report results on different configurations in table \ref{tbl:optimization_methods}. The following configurations have been tested:(i) using a single representation (only $Z$), (ii) when using only the clustering loss, (iii) the results obtained using an ADMM optimization scheme (without optimizing the parameters per dataset).
For (i) and (ii), we try various parameter settings (using a grid of possible parameter configurations) and report the best results.

\subsection{Semi-Supervised Clustering Evaluation}
\begin{table}[]
\centering
\setlength{\tabcolsep}{0.5em}
\begin{tabular}{l|cc|cc|cc}
\multirow{2}{*}{\begin{tabular}[c]{@{}l@{}}\#labeled \\ connections\end{tabular}} & \multicolumn{2}{c|}{USPS} & \multicolumn{2}{c|}{FRGC} & \multicolumn{2}{c}{CMU-PIE} \\ \cline{2-7} 
                                                                                  & NMI         & ACC        & NMI         & ACC         & NMI           & ACC          \\ \hline
0                                                                                 & 0.78       & 0.58      & 0.74       & 0.48       & 0.87         & 0.63        \\
0.1k                                                                               & 0.78       & 0.57      & 0.74       & 0.50       & 0.87          & 0.64        \\
1k                                                                              & 0.85       & 0.76      & 0.77       & 0.53       & 0.90         & 0.73        \\
5k                                                                              & 0.92       & 0.86      & 0.82       & 0.57        & 0.93         & 0.79       
\end{tabular}
\caption{Semi-supervised clustering results. Above, we report the clustering performance as the number of labeled MKNN connections increases. }
\label{tbl:semi_supervised_results_ADMM}
\end{table}

As detailed in Section \ref{sec:semi_method}, our method naturally extends to a semi-supervised framework. 
In Table \ref{tbl:semi_supervised_results_ADMM}, we demonstrate the improvement in the clustering performance as the number of labeled MKNN connections increases on three different datasets. 
To put matters in perspective, the percentage of labeled connections out of the entire dataset is less than 0.2\% for all the results reported in the table. 
As the table illustrates, a small portion of labeled connections can significantly boost performance.

\subsection{3D Point Cloud Clustering}
To demonstrate the robustness of our method, we further extend it to work on datasets of 3d objects. Contrary to the datasets we have shown so far, the feature representation of the point clouds has to be permutation-invariant and the reconstruction should match the shape outline and not the exact point coordinates. Therefore, a different autoencoder architecture and loss need to be used. 
We use the architecture in \cite{achlioptas2017representation} with a Chamfer loss instead of the fully connected architecture. We perform two sets of experiments on 3d data -- inter-class clustering, where the dataset contains different classes of 3d objects, and intra-class clustering, where the dataset contains sub-categories of the same class.

For these experiments, we use objects from ShapeNet \citet{chang2015shapenet}, which are sampled to create point clouds with 2048 points. The autoencoder is first trained for 1000 iterations using an Adam optimizer with a learning rate of 0.0005. During the clustering stage the autoencoder learning rate is set to 0.0001 and the learning rate of $U$ is set to 0.0001. The number of epochs between $\mu$ update is set to 30.

\subsubsection{Inter-class Clustering}
\begin{table}[]
\centering
\setlength{\tabcolsep}{0.5em}
\begin{tabular}{l|cc}
method &  NMI         & ACC		\\ \hline
AP                                                                              & 0.59       & 0.16         \\
HDBSCAN                                                                              & 0.52       & 0.50       \\
CPAC                                                                              & 0.64       & 0.56       \\ \hline
0.1k                                                                               & 0.70       & 0.65        \\
1k                                                                              & 0.73       & 0.65         \\
5k                                                                              & 0.77       & 0.73       
\end{tabular}
\caption{Clustering results on ShapeNet10 point cloud objects. In the first three rows, we compare the clustering results of different non-parametric algorithms on the initial embedded space to our algorithm. In the three bottom rows, we report the performance as the number of labeled MKNN connections increases. }
\label{tbl:semi_supervised_point_cloud}
\end{table}

We show clustering results on the ModelNet10 dataset \cite{wu20153d}. We visualize the shape embedding at the beginning (after the initialization of the autoencoder) and at the end of the clustering stage. Some of the classes in ModelNet10 contain a variety of shapes, for example, the chairs class, as described further in Section \ref{sec:intra-class}). Furthermore, some of the classes contain very similar shapes, like the dresser and the night-stand classes. Therefore, it is not surprising that in the final embedding some of the classes are clearly separated, while some are embedded closer. In table \ref{tbl:semi_supervised_point_cloud}, we report results using different  clustering algorithms on the initial latent embedding. There is no comparison to other deep clustering algorithms since their architecture is not suitable for point clouds.
We further show the improvement of clustering results when using the extension to semi-supervised clustering (in Table \ref{tbl:semi_supervised_point_cloud}, below the line).
\begin{figure*}
\includegraphics[width=\textwidth]{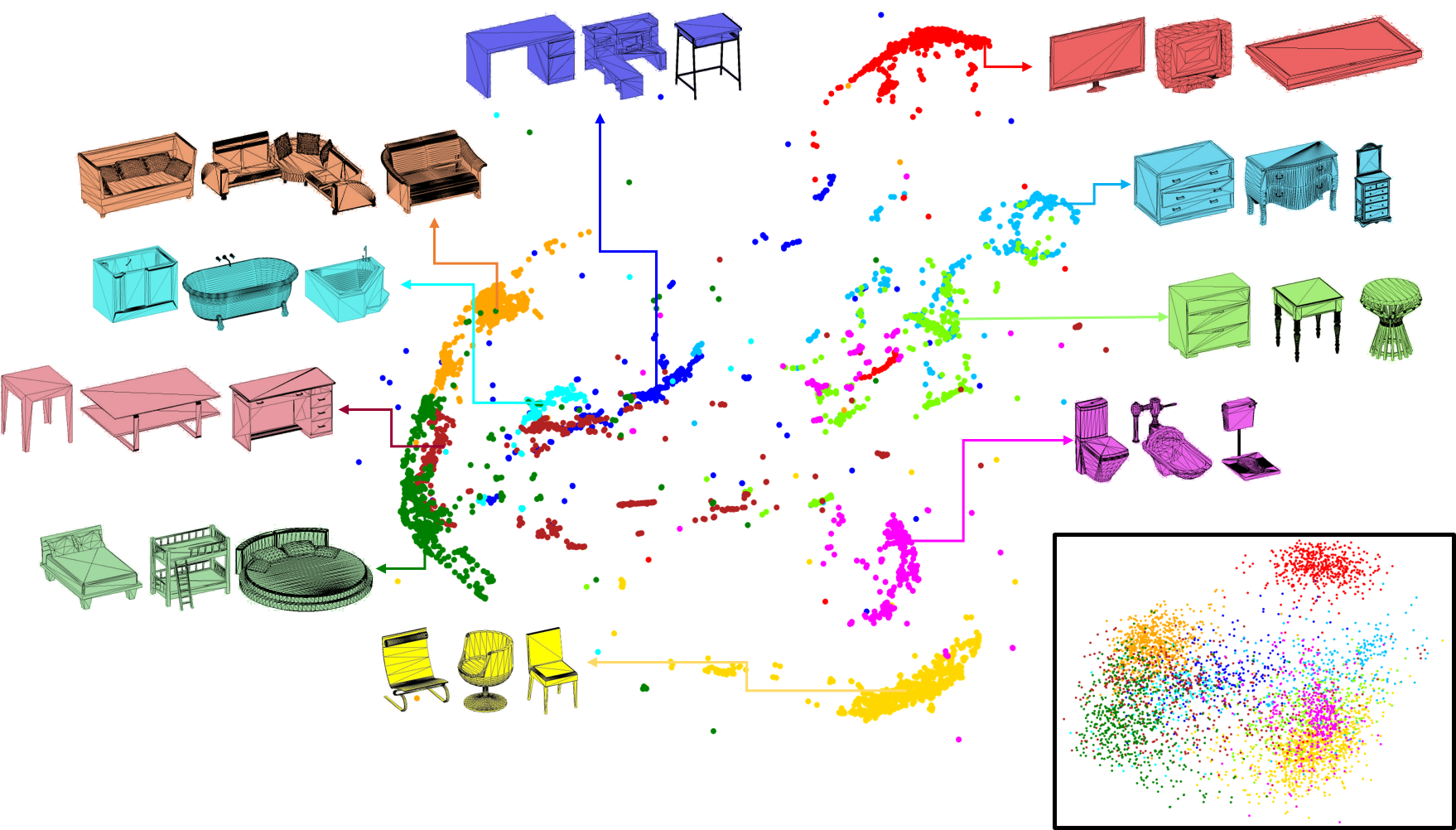}
\caption{PCA visualization of the embedding of 10 shape classes. Above, we visualize the embeddings after the initial training of the autoencoder (on the bottom right) and after running our clustering algorithm. The points are colored according to their ground-truth class and shape representatives of each class are also illustrated.}
\label{fig:point cloud pca}
\end{figure*}

\subsubsection{Intra-class Clustering}
\label{sec:intra-class}
\begin{figure*}
\includegraphics[width=\textwidth]{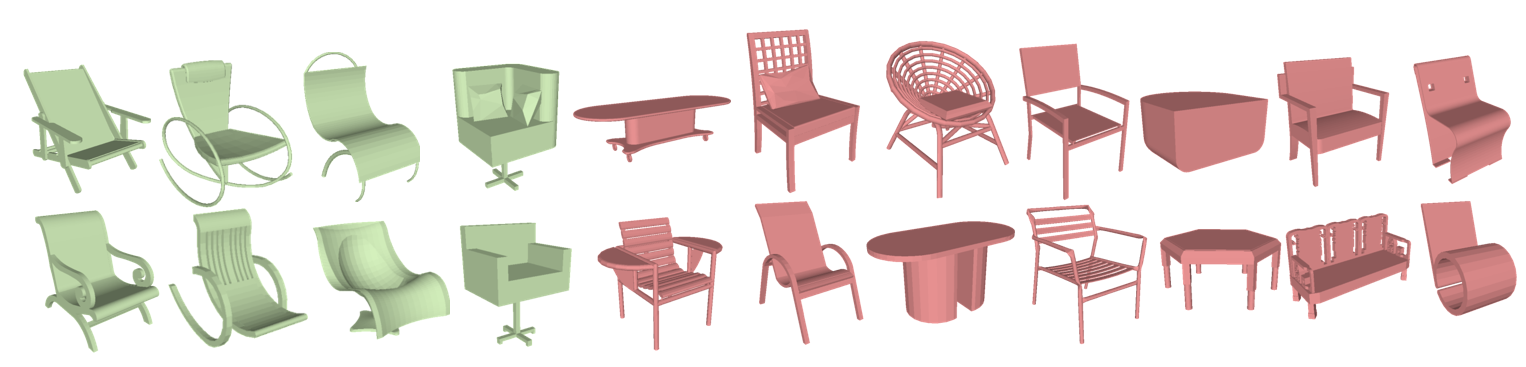}
\caption{Examples of shape pairs with a high clustering loss. The pairs are colored according to the labeling of the user. On the left are the must-link pairs (colored in green), and on the right are the cannot-link pairs (colored in red).}
\label{fig:chair pairs}
\end{figure*}

\begin{figure*}[htp]
\jsubfig{\includegraphics[height=4.3cm]{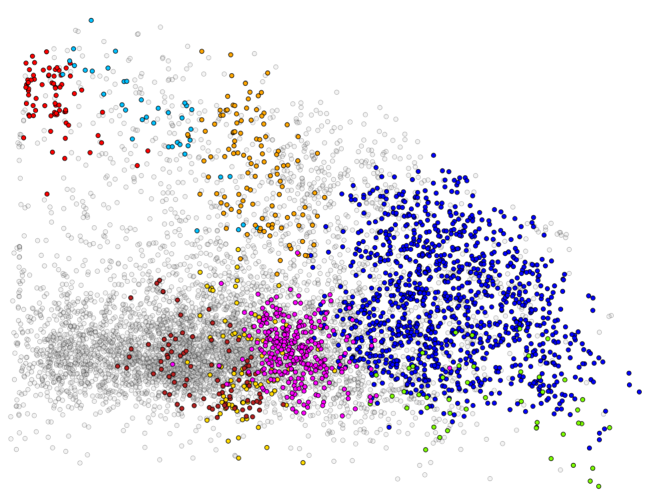}}
	{(a)}%
    \hfill
 \jsubfig{\includegraphics[height=4.3cm]{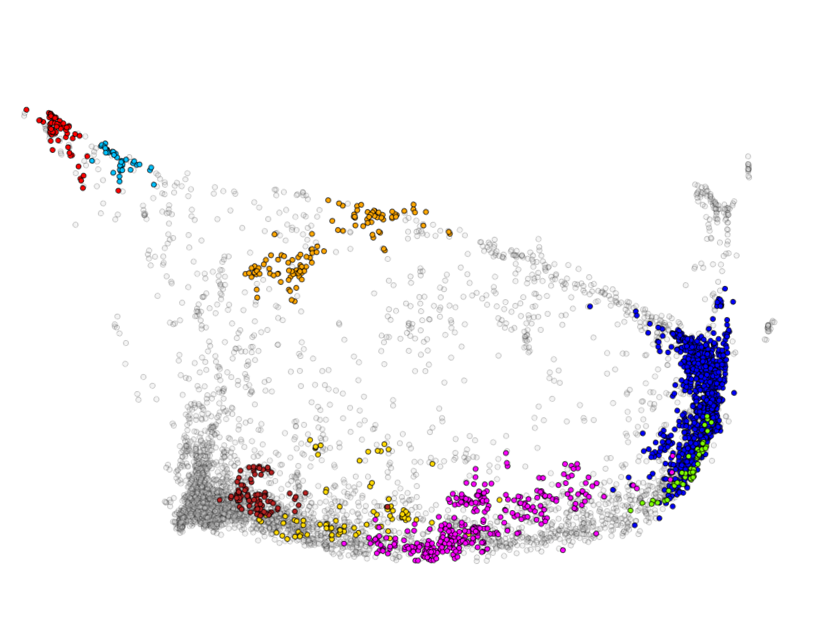}}
	{(b)}%
    \hfill
  \jsubfig{\includegraphics[height=4.3cm]{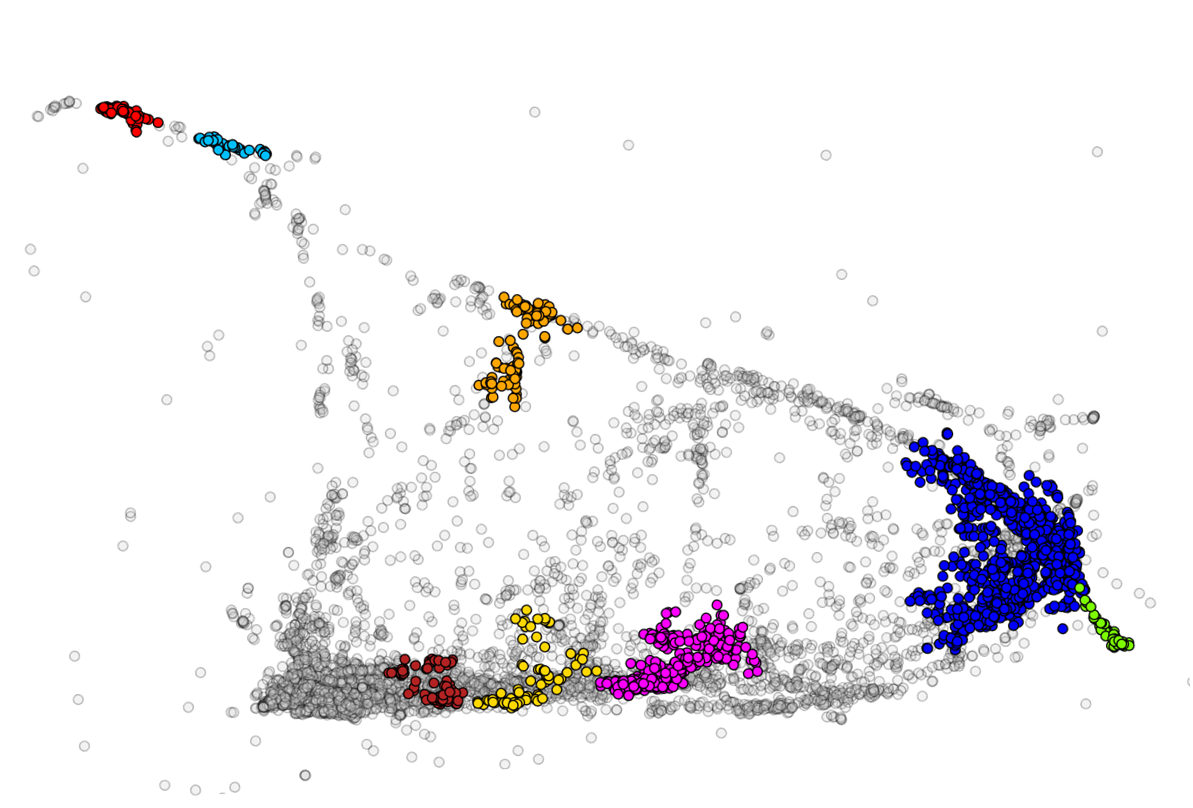}}
	{(c)}%
\caption{PCA embedding of the clustering representation of 3D shapes (a) after the initial training of the autoencoder, (b) after the clustering stage, and (c) after labeling 1000 connections. We randomly select a few clusters from (c) and illustrate the distribution of these sampled points in the previous two stages. As the figure illustrates, the shape points are more visually separable in (c). In figure \ref{fig:teaser}, we illustrate a few representative chairs from each cluster. }
\label{fig:chairs clustering}
\end{figure*}

We demonstrate our clustering results on the 6778 chairs available on ShapeNet. This class is not only large in size, but more importantly, it is highly variable and contains objects that can clearly be separated into sub-categories.
As are no ground truth labels for this experiment, we demonstrate the improvement in the clustering representation and the clustering results in a qualitative manner, by visualizing the embedding of the chairs before and after the clustering stage, and also after labeling 1000 pairwise connections. See Figure \ref{fig:chair pairs} for a visualization of the labeled pairs, and Figure \ref{fig:chairs clustering} for a visualization of the embeddings. As Figure \ref{fig:chairs clustering} illustrates, the embedding of the data becomes more tightly clustered and the separation of the chairs into different clusters becomes more meaningful.

\section{Conclusions}
We presented a clustering technique which leverages the strength of deep neural network. Our technique has two unique properties: (i) first, unlike previous methods, it is non-parametric and does not assume any knowledge about the expected number of clusters, and (ii) the technique builds on pairs of points rather than on centroids. This pairs-based approach allows the integration of additional supervised pairs which enhances the clustering performance. The supervision can be provided actively by a user, or from a small subset of labeled data. 
Technically, we constructed a Siamese autoencoder which learns a representation that is applicable for clustering complex structures. 
To boost performance, we turned our optimization problem into a constrained one, which creates a separation between the clustering and the reconstruction losses. We showed how this separation can be further enhanced by incorporating an ADMM optimization scheme into the system.

We demonstrated that our approach obtains state-of-the-art results on various datasets, including 2D images and 3D shapes, in an unsupervised setting where the class labels are not provided. Furthermore, we demonstrated that when a small portion of the labels are provided, or are manually annotated using pairwise connections, our approach naturally extends to incorporate these constraints. We presented an approach based on analyzing the losses for carefully selecting these additional constraints and demonstrated a significant boosting of performance using just a small percentage of annotated data.

We should stress again, that we deliberately avoided fine-tuning our technique for the different datasets. 
We believe that with moderate fine tuning per dataset, our performance can be further improved. 

In our presented method we employ a MKNN neighborhood connectivity to automatically extract conservative pairs that derive the clustering. The choice of MKNN is advantageous since the number of false positives is relatively low, and, using semi-supervision, we are able to further overcome this issue. Nonetheless, in the future, we would like to explore possibly a more complex neighborhood connectivity that considers local density distributions. In general, we believe that combining our pairs-driven technique with a density-centered approach is an interesting avenue for future research.


\bibliographystyle{ACM-Reference-Format}
\bibliography{clustering}
\end{document}